\documentclass[10pt,twocolumn,letterpaper]{article}

\usepackage{cvpr}              

\usepackage[dvipsnames]{xcolor}

\definecolor{cvprblue}{rgb}{0.21,0.49,0.74}
\usepackage[pagebackref,breaklinks,colorlinks,citecolor=cvprblue]{hyperref}

\def\paperID{*****} 
\def\confName{CVPR}
\def\confYear{2024}

\title{FAPNet: An Effective Frequency Adaptive Point-based Eye Tracker}
\author{Xiaopeng Lin \thanks{equal contribution. \dag corresponding author.}, Hongwei Ren\textsuperscript{*}, Bojun Cheng \dag\\
The Hong Kong University of Science and Technology (Guangzhou)\\
{\tt\small\{xlin746,hren066\}@connect.hkust-gz.edu.cn, bocheng@hkust-gz.edu.cn}
}

\begin{document}
\maketitle
\begin{abstract}
Eye tracking is crucial for human-computer interaction in different domains. Conventional cameras encounter challenges such as power consumption and image quality during different eye movements, prompting the need for advanced solutions with ultra-fast, low-power, and accurate eye trackers. Event cameras, fundamentally designed to capture information about moving objects, exhibit low power consumption and high temporal resolution. This positions them as an alternative to traditional cameras in the realm of eye tracking. Nevertheless, existing event-based eye tracking networks neglect the pivotal sparse and fine-grained temporal information in events, resulting in unsatisfactory performance.  Moreover, the energy-efficient features are further compromised by the use of excessively complex models, hindering efficient deployment on edge devices. In this paper, we utilize Point Cloud as the event representation to harness the high temporal resolution and sparse characteristics of events in eye tracking tasks. We rethink the point-based architecture PEPNet with preprocessing the long-term relationships between samples, leading to the innovative design of FAPNet. A frequency adaptive mechanism is designed to realize adaptive tracking according to the speed of the pupil movement and the Inter Sample LSTM module is introduced to utilize the temporal correlation between samples. In the Event-based Eye Tracking Challenge, we utilize vanilla PEPNet, which is the former work to achieve the $p_{10}$ accuracy of 97.95\%. 
On the SEET synthetic dataset, FAPNet can achieve state-of-the-art while consuming merely 10\% of the PEPNet's computational resources. Notably, the computational demand of FAPNet is independent of the sensor's spatial resolution, enhancing its applicability on resource-limited edge devices. 
\end{abstract}    
\section{Introduction}
\label{sec:intro}
Eye tracking stands as a pivotal technology within the realm of human-computer interaction and finds extensive applications in augmented reality (AR), extended reality (XR), medicine, and psychology \cite{guenter2012foveated, konrad2020gaze}. It involves locating the pupil position and capturing the eye movement trajectory. Initially, the human eye undergoes rapid movements during specific motions, often surpassing speeds of $300^\circ/s$ \cite{ramachandran2012encyclopedia}. This observation necessitates the adoption of very fast sampling, typically in the kHz range, to effectively capture minute yet swift eye movements as they commence and trace their trajectories. Additionally, it is noteworthy that the eye's acceleration frequently attains astronomical values, exemplified by instances such as $24,000^{\circ}/s^{2}$ \cite{angelopoulos2021event}. This highlights the remarkable capability of eye muscles to execute forceful and high-frequency movements. Hence, a high-speed and precise eye-tracking system has become an essential requisite.

High-speed traditional camera-based eye tracking systems are expensive and power-intensive. The motions captured by cameras in the near-eye scenarios are redundant, as the primary motion typically originates from the pupil region while the remaining areas are static \cite{chen20233et}. Consequently, the main challenge in the eye tracking task is the efficient capture of motion information and the development of lightweight model to enhance the overall efficiency and effectiveness of eye-tracking systems in mobile devices \cite{yang2021edge}.

Event-based cameras, such as Dynamic Vision Sensors (DVS) \cite{lichtensteiner2008128x128}, operate fundamentally differently from traditional cameras. Instead of synchronously capturing entire image frames at a fixed frame rate, each pixel in event cameras independently triggers an event when local brightness changes reach a certain threshold. This mechanism allows event cameras to operate with extremely low power consumption while offering high dynamic range and temporal resolution with negligible latency \cite{rebecq2019high}. Due to the independent operation of each pixel, event cameras can respond instantly to changes, circumventing the frame rate \cite{zhang2022unifying} limitations and motion blur \cite{jiang2020learning} inherent to conventional video cameras.

The data captured by event cameras exhibit intrinsic sparsity, which is highly advantageous for eye tracking applications \cite{chen20233et, stoffregen2022event, zhao2024ev}. Besides, event cameras excel in detecting subtle changes in eye movement, capturing only the dynamic elements within the visual field, primarily the pupil movements. This sparsity ensures that the system focuses on the essential aspects of ocular activity, significantly enhancing the quality of data for eye-tracking. The asynchronous nature of event cameras allows for high temporal resolution and immediate detection of rapid eye movements, making them exceptionally suited for sophisticated eye-tracking tasks that require precision and speed. This capability to selectively capture relevant ocular events minimizes data volume and processing load, leading to efficient and accurate tracking of eye movement in various motion conditions.


However, applying event data to eye tracking tasks presents several challenges. The first challenge is how to efficiently represent events to leverage the rich motion information of event data, which highly corresponds to the pupil movement. Frame-based methods struggle to accurately capture motion details in event sequences with low frame rates, especially during rapid pupil movements. The derived pupil locations represent the average position of the pupil within a frame, leading to suboptimal eye tracking performance. Besides, both voxel-based and frame-based event representations require a balance between computational load and spatial resolution, necessitating a trade-off between accuracy and efficiency. The second challenge is how to design lightweight neural networks capable of performing high-accuracy eye tracking on edge devices in the near-eye tracking scenario. To deal with these challenges, we introduce the lightweight Frequency Adaptive Point-based Eye Tracking Network (FAPNet) for eye tracking based on our former work PEPNet \cite{ren2024simple}. The raw Point Cloud is directly leveraged as the network input to fully utilize the high temporal resolution and inherent sparsity of events. A point-based network is designed to aggregate the spatial and temporal information in a lightweight formula. Additionally, the parameter and floating-point operations (FLOPs) in our model remain constant regardless of the input camera's resolution, in contrast to frame-based methods \cite{chen20233et}. The main contributions can be concluded as follows:
\begin{itemize}
    \item We first introduce the Point Cloud representation into the eye-tracking field to fully utilize the fine-grained temporal information of the events highly corresponding to pupil movement.
    \item We propose a lightweight model consisting of local-global spatial information aggregation and the long-short temporal information correlation. To achieve high accuracy tracking while maintaining low power consumption.  
    \item We design a frequency adaptive mechanism that adjusts the tracking rate based on the velocity of pupil movement, enhancing accuracy while reducing computational load.
\end{itemize}

\section{Related Work}
\label{sec:formatting}

\subsection{Event Representation}

Event camera has been widely utilized in many applications, such as action recognition  \cite{innocenti2021temporal}, object classification \cite{bi2020graph,huang2024clif}, image deblurring  \cite{zhang2022unifying}, 3D reconstruction \cite{chamorro2022event} and so on. To efficiently integrate the asynchronous and sparse events into the conventional neural network, event streams are transformed into specific representations as can be concluded into frame-based \cite{krafka2016eye, bardow2016simultaneous}, voxel-based  \cite{zihao2018unsupervised} and point-based \cite{wang2019space, ren2023ttpoint} representations. Event Frame is the most widespread due to its compatibility with conventional computer vision algorithms in various applications \cite{planamente2021da4event}. The voxel-based representations accumulate the events in their pixel locations and the temporal dimension is cut into temporal bins to preserve the coarse temporal information inherent in the events. The point-based representations regard the events as the Point Cloud by treating the $t$ to $z$ \cite{ren2023spikepoint}, which can better utilize the sparsity of the events and preserve the fine-grained temporal information\cite{ren2024rethinking}. Compared to other representations, Point Clouds do not require data format transformation and can directly process the raw data generated by the event camera, eliminating the need for event-free region calculations.

\subsection{Point Cloud Networks}
PointNet \cite{qi2017pointnet} revolutionizes Point Cloud processing paradigms by directly handling Point Cloud input. Building upon this, PointNet++ \cite{qi2017pointnet++} enhances its capabilities through its integration of the Set Abstraction (SA) module, allowing for both global and local analyses within the hierarchical structure of Point Clouds. Instead of utilizing the simple multilayer perceptron as a feature extractor, PointConv \cite{wu2019pointconv} employs deep neural networks to process Point Clouds, PointTransformer \cite{zhao2021point} leverages Transformer as the backbone network, and more effective networks are proposed \cite{ma2022rethinking,qian2022pointnext,xiao2023unsupervised}. Event Cloud and Point Cloud exhibit remarkable similarities in their representations, and sophisticated Point Cloud networks have shown promising performance in event-based as well.
STNet \cite{wang2019space} processes the temporal information without loss of representation with x and y. They applied PointNet++ to complete the task of gesture recognition. PAT \cite{yang2019modeling} incorporates self-attention and Gumbel subset sampling to improve the performance in recognition tasks. TTPOINT \cite{ren2023ttpoint} is a lightweight and generalized Point Cloud network that achieves impressive results in action recognition tasks. SpikePoint \cite{ren2023spikepoint} designs a point-based spiking neural network architecture, which excels at processing sparse event cloud data with few parameters and maintains low power consumption. Furthermore, PEPNet \cite{ren2024simple} demonstrated outstanding performance in the regression task of event camera pose relocalization, outperforming all conventional frame-based methods.

\subsection{Eye Tracking}

Current camera-based solutions of eye tracking can be divided into two parts depending on the sensors utilized: conventional eye tracking \cite{morimoto2005eye} and event-based eye tracking \cite{angelopoulos2021event}. Conventional eye tracking receives the full-resolution eye images as input, which consists of model-based methods \cite{wang2017real, guestrin2006general} and appearance-based methods \cite{lu2014adaptive, wood2016learning}. Model-based methods predict pupil location through the eyes' mechanical modeling and gaze estimation. The appearance-based methods leverage neural networks to learn a mapping from an eye image to gaze directions. Compared with the appearance-based methods, the model-based methods are regarded to provide better tracking accuracy \cite{fengreal}.

Event-based eye tracking methods leverage the inherent attribution of event data to realize low-latency tracking. Ryan et al. \cite{ryan2021real} propose a fully convolutional recurrent YOLO architecture to detect and track faces and eyes. This method transforms the raw event streams into the voxel grids as the input representation. It achieves impressive detection accuracy in eye location detection instead of near-eye pupil tracking. Stoffregen et al. \cite{stoffregen2022event} design Coded Differential Lighting method in corneal glint detection for eye tracking and gaze estimation. It can realize a kHz sampling rate tracking. Angelopoulos et al. \cite{angelopoulos2021event} propose a hybrid frame-event-based near-eye gaze tracking system, which can offer update rates beyond 10kHz tracking. Zhao et al. \cite{zhao2024ev} also leverage the near-eye grayscale images with event data for a hybrid eye tracking method. The U-Net architecture is utilized for eye segmentation and a post-process is designed for pupil tracking. Chen et al. \cite{chen20233et} presents a sparse Change-Based Convolutional Long Short-Term Memory model to achieve high tracking speed in the synthetic event dataset. 


Although the existing event-based methods have achieved impressive tracking results, there still exist some limitations.
Both voxel and event frame representations are derived from the original events, and data format transformation consumes significant resources and time. These representations also lack fine-grained spatial-temporal information and frequency adaptive tracking for different pupil states. In contrast to prior methods, we directly leverage the raw event cloud as the network input to fully utilize the inherent high temporal resolution attribution of events. And the point-based network is also designed to realize frequency-adaptive tracking according to different pupil states in a lightweight design principle.
\section{Real and Synthetic Datasets}
\label{sec:Data}

\begin{figure}[ht]
    \centering
    \includegraphics[width=1\linewidth]{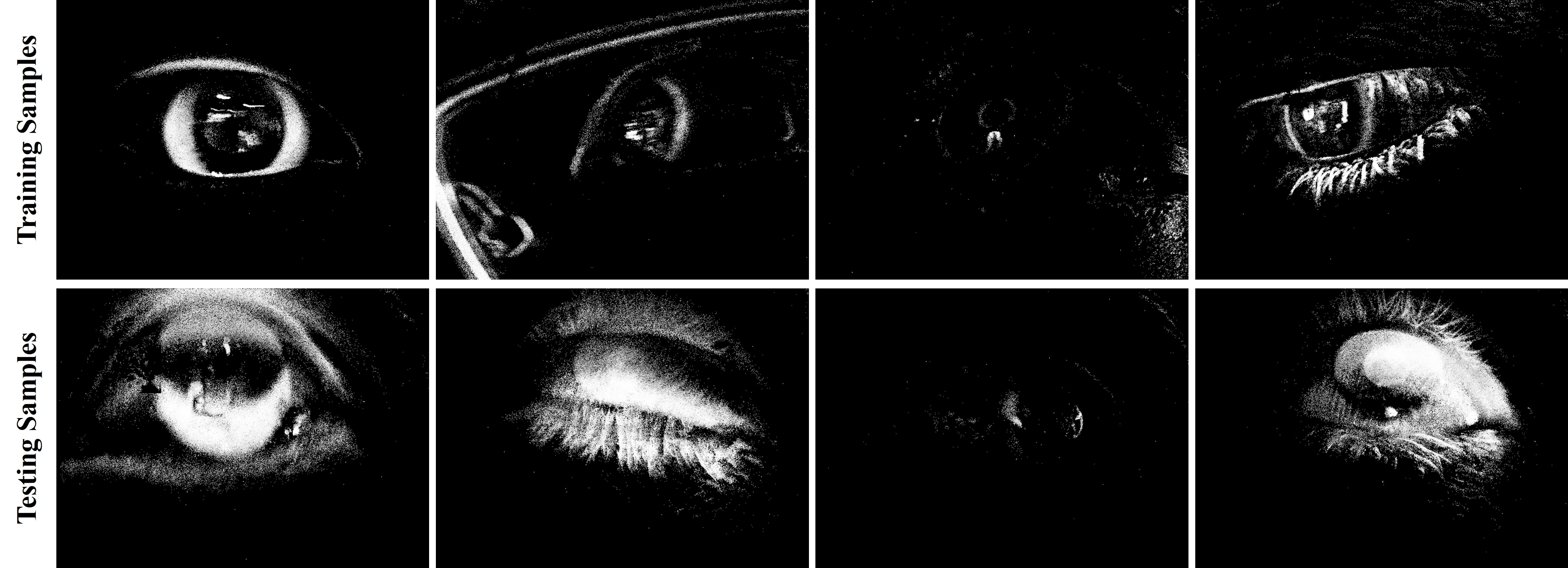}
    \caption{Real Data overview. The first row is the selected training samples and the second row is the selected testing samples. All the visualized samples are generated by stacking the events in a 50ms time interval.}
    \label{fig:dataover}
\end{figure} 

\begin{figure}[ht]
    \centering
    \includegraphics[width=1\linewidth]{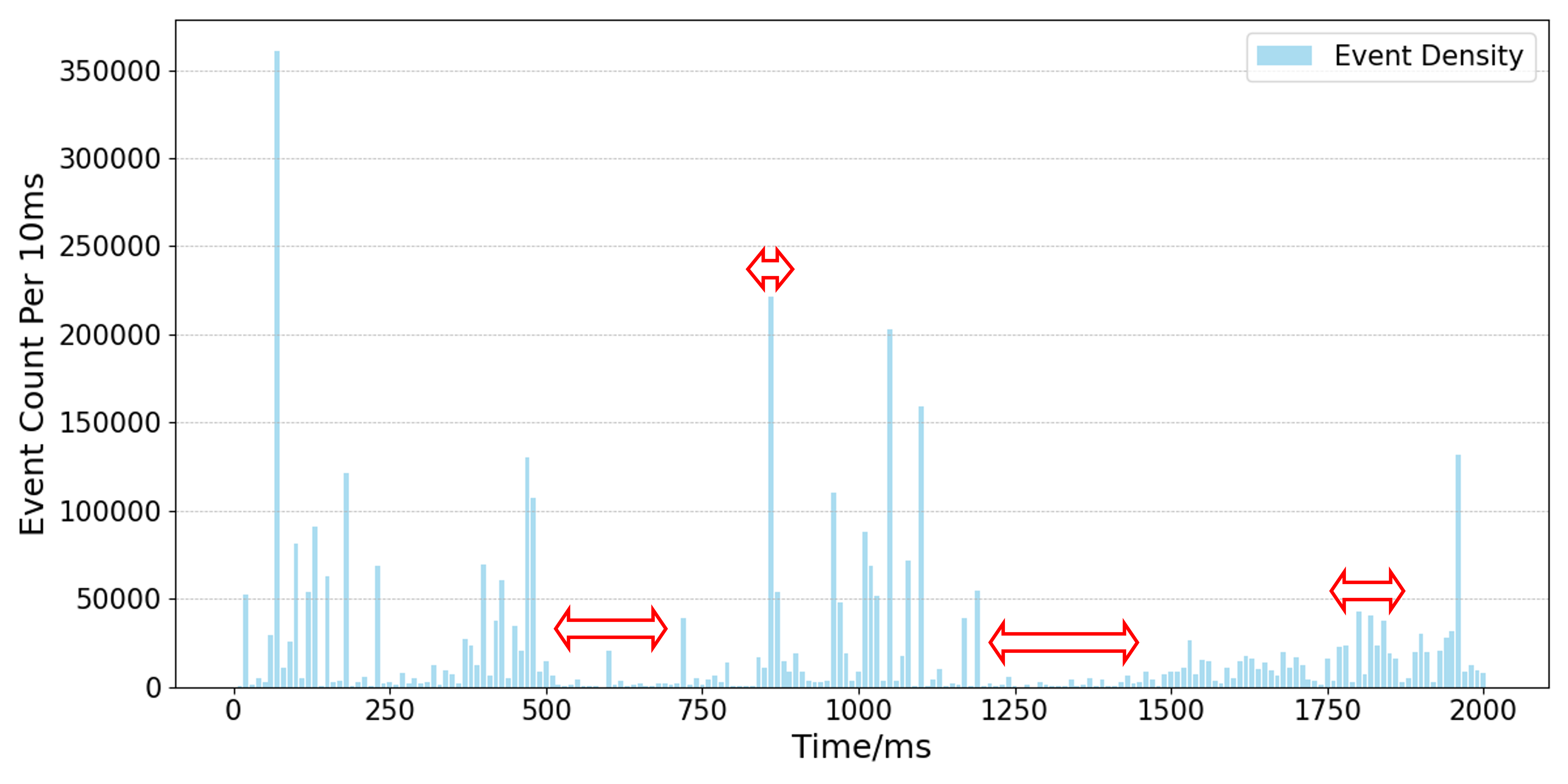}
    \caption{Event density. The x-axis is the time in ms and the y-axis is the event count of each 10ms sample. The red arrow is the dynamic window length of each sample.}
    \label{fig:density}
\end{figure}

\subsection{Data Overview}

\textbf{Real data in Challenge.} The real event dataset EET+ \cite{wang2024ais_event} is sourced from the Event-based Eye Tracking Challenge hosted at the AI for streaming Workshop CVPR 2024. The dataset contains 40 motions for training and 12 motions for testing, each with a resolution of 640 $\times$ 480. The training motion sequences are annotated with pupil locations at a 100Hz frequency, with an evaluation focus on 20Hz accuracy. The main purpose of the model is to predict the location of the pupil in 20Hz tracking frequency. 

\textbf{Synthetic dataset.} SEET \cite{chen20233et} is utilized for the synthetic event data evaluation. SEET is simulated from an RGB dataset, which is named Labeled Pupils in the Wild. Event streams are generated through the v2e simulator with the $240 \times 180$ resolution. For a fair comparison, we follow the same dataset setting as the original paper.

The two datasets have rich diversity with various eye shapes and pitch angles of users. There exist some factors which easily affect the prediction of the pupil location, such as the eyelashes, closed eye state, and glasses as can be seen in \cref{fig:dataover}. The event density which means the event count of each 10ms sample also varies in each motion as shown in \cref{fig:density}, which is positively correlated to the frequency of pupil movement. The sample with few events indicates that the pupil is in a stationary state during the time interval. It is efficient to develop a robust algorithm with frequency adaptive to focus more on the pupil's fast or slow movement process.

\begin{figure}[ht]
  \centering
   \includegraphics[width=1\linewidth]{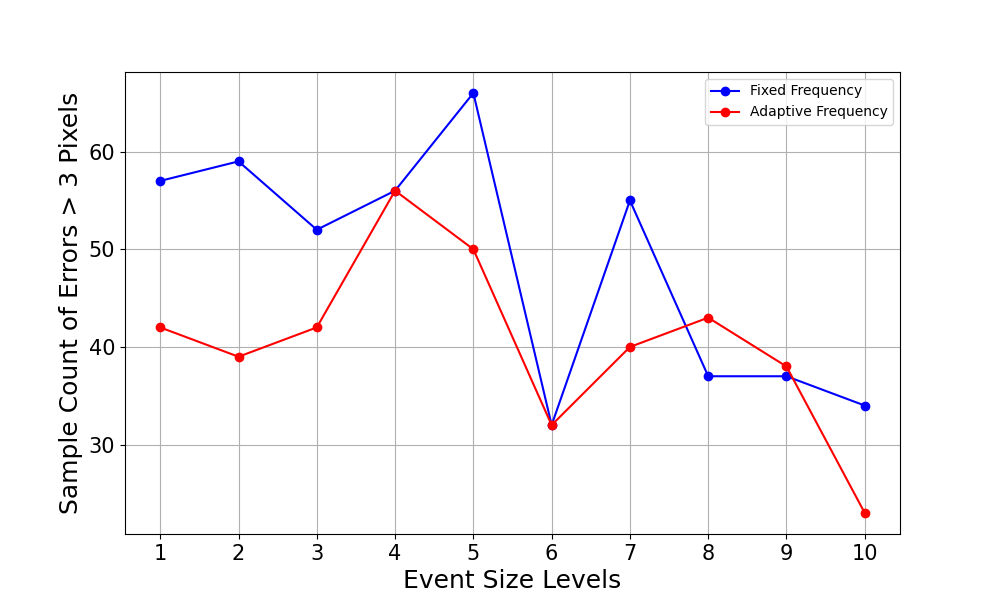}

   \caption{Correlation between event density and errors. The X axis is the event size level and the Y axis is the number of samples with errors above 3 pixels in each level. The higher the event size level, the larger the event numbers in each sample. The blue line is the fixed frequency eye tracking and the red line is the adaptive frequency tracking.}
   \label{fig:errorwithnum}
\end{figure}

\subsection{Data Processing}

The data processing of this paper consists of three parts: Sliding Window Split, Frequency Adaptive Window Expanding, and Trajectory Augmentation.

\textbf{Sliding Window Split.} To fully utilize the high temporal resolution of event data and the annotated information, the training data processing involves split motions into 10ms samples with a non-overlapping 10ms sliding window. Each sample correlates with a 100Hz label. 

\begin{figure*}[ht]
  \centering
   \includegraphics[width=0.9\linewidth]{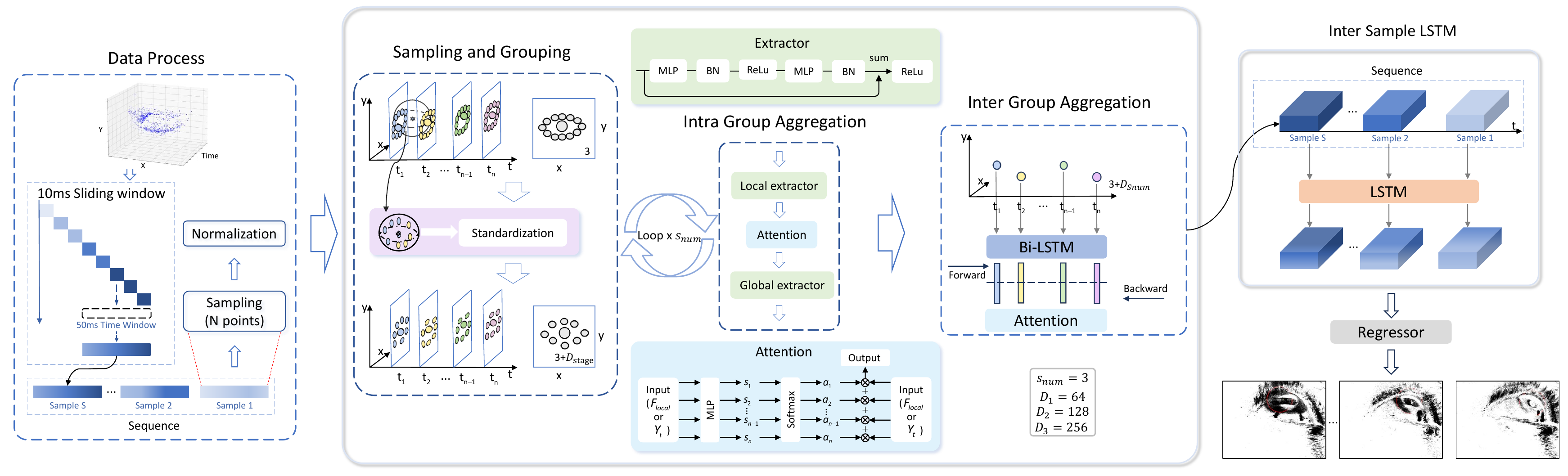}

   \caption{FAPNet's Network Architecture. The input Event Cloud is directly processed using a sliding window, sampling and normalization, eliminating the necessity for any format conversion. Each of the $S$ samples is fed into the network as a sequence. The input passes through $S_{sum}$ loops between the Sampling and Grouping module and the Intra Group Aggregation module for spatial feature abstraction and extraction. The input is then passed through a bidirectional LSTM to extract temporal features. The Inter Sample LSTM Module is designed to aggregate information within the sequence by the LTSM, culminating in a regressor responsible for eye tracking.}
   \label{fig:Network}
\end{figure*}

\textbf{Frequency Adaptive Window Expanding.} 
In this paper, we design the frequency adaptive window to adaptively adjust the tracking frequency according to the event density. Specifically, if event counts within the 10ms time window exceed a predetermined threshold, the window size remains constant. Otherwise, the window will gradually expand equally to both sides until its event number reaches the threshold or a maximum window length (eg. 100 ms). As shown in \cref{fig:density}, the time window length is small in the time periods with lots of events. While it is large when the events are limited.

As illustrated in \cref{fig:errorwithnum}, we conduct an analysis of error relative to event density. All sliding windows are categorized into 10 distinct sub-bins based on the event density. Analysis of the fixed sliding window (blue) reveals a significant $p_3$ error in the leftmost bin, which has the fewest events. This low event count, indicative of slower pupil movement, suggests the need to expand the sliding window to reduce error. By employing an adaptive sliding window (red), the error is significantly reduced, particularly in windows with low event densities.

\textbf{Trajectory Augmentation.} Eye movement trajectories are various in real world applications, such VR and AR. To enrich the data variety and trajectory complexity, we apply data augmentation techniques, such as trajectory inversion to avoid the model over-fitting to specific trajectories. These steps are pivotal in optimizing the model's learning efficiency across diverse motion trajectories. It can give the model stronger generalization ability, to better carry out efficient and accurate eye tracking in practical applications.

\section{Method}
\label{sec:Method}
\subsection{Preprocessing}
The event data needs to undergo downsampling from the packaged samples and perform normalization operations before feeding into the network. In this study, we randomly select $N$ points from each sample to unify the number of points, and each of the $S$ samples is fed into the network as a sequence. The normalization operation is conducted through the spatial and temporal dimensions, as shown in Eq. \ref{eq:spatial_normalization}. In the spatial dimension, the coordinates are normalized by the resolution $w$ and $h$ of the input. The temporal dimension normalization is conducted by subtracting the smallest timestamp of each sample and dividing it by the timestamp length between the start and end events. 
\begin{equation}
S_i^{'}= \left( \frac{x}{w}, \frac{y}{h} \right), \quad T_i^{'} = \frac{T_i - \min(T)}{\max(T) - \min(T)}
\label{eq:spatial_normalization}
\end{equation}
where ${S_{i}}^{'}$ is the spatially normalized coordinate of a event in the $i_{\text{th}}$ sample, $x$ and $y$ are the original coordinates, $T_i$ represents the original timestamp of points in the $i_{\text{th}}$ sample.

\subsection{Network Architecture}
As shown in \cref{fig:Network}, FAPNet consists of four main modules: Sampling and Grouping module, Intra Group Aggregation module, Inter Group Aggregation module, and Inter Sample LSTM module. After going through these modules, a regressor predicts the pupil's location as the tracking result. The lightweight asynchronous architecture enables the model to achieve high-accuracy tracking with low power consumption which is suitable for edge devices.

\subsubsection{Sampling and Grouping} 
After preprocessing, the data sequence will be fed into the Sampling and Grouping module to capture essential spatial and temporal information. The Farthest Point Sampling (FPS) and K-nearest neighbors (KNN) are leveraged to extract local spatial information and temporal information efficiently. The standardization operation is also applied in each group to ensure consistent variability between points within the group.

\subsubsection{Intra Group Aggregation} 
To enhance feature extraction and integration, an Intra Group Aggregation module is designed to extract the local spatial and local temporal features. The feature extraction is conducted through the extractor, which consists of a simple Multi-Layer Perceptron (MLP) with a residual connection. The information integration within each group is achieved through the attention mechanism. The Intra Group Aggregation module combines the extractors and the attention mechanism to distill and consolidate local spatial-temporal information efficiently. 

\subsubsection{Inter Group Aggregation}

The local temporal features extracted from the Intra Group Aggregation module are independent and parallel between different groups. We also employ an Inter Group Aggregation module to integrate the global temporal information among groups along the timestamp dimension. A bidirectional LSTM network synergized with an attention mechanism is proposed to elucidate temporal relationships across different event groups. The Inter Group Aggregation module with a lightweight design facilitates precise pupil position regression, which is important for realizing low power consumption with high accuracy in the eye tracking task.

\subsubsection{Inter Sample LSTM} 
To better process and analyze the time-sequential event samples, we design the Inter Sample LSTM module to capture the long temporal dependencies of the sequential samples and each sequence comprises $S$ samples. The long temporal correlation is conducted as follows:  
\begin{align}
    \mathbf{h}_t = f(&\mathbf{W}_h \cdot \mathbf{x}_t + \mathbf{U}_h \cdot \mathbf{h}_{t-1} + \mathbf{b}_h), \\
    &\mathbf{y}_t =\mathbf{V} \cdot \mathbf{h}_t + \mathbf{b}_y, \quad
\end{align}
where $\mathbf{W}_h$, $\mathbf{U}_h$, and $\mathbf{b}_h$ represent the weight matrix and bias vector of the Inter Sample LSTM, $\mathbf{V}$ and $\mathbf{b}_y$ stand for the weight matrix and bias vector of its output layer. $f(\cdot)$ is the activation function, which can be chosen as sigmoid or other functions. $x_t$ is the Point Cloud feature of the sample at the time t. Finally, $y_t$ is utilized as input to the regressor for coordinate regression. 

Combining the group-level short temporal correlation and the sample-level long temporal dependency, the features of these samples are integrated to predict the corresponding $S$ pupil locations. The temporal connections are revealed between sequential samples, which significantly enhance the accuracy and efficiency of eye tracking.
\begin{table*}[tb]
\centering
\renewcommand\arraystretch{1.2}
\scalebox{0.8}{
\begin{tabular}{ccccccccc}
\hline
Method          & Resolution & Representation & Param (M)         & FLOPs (M)      & $p_{3}$             & $p_{5}$             & $p_{10}$            & mse(px)       \\ \hline
\textbf{FAPNet} & 180 x 240  & P     & \textcolor[rgb]{0,0,1}{0.29} & 58.7 & \textcolor[rgb]{1,0,0}{0.920} & \textcolor[rgb]{1,0,0}{0.991} & 0.996 & \textcolor[rgb]{1,0,0}{1.56} \\
PEPNet\cite{ren2024simple}          & 180 x 240  & P              & 0.64          & 443           & \textcolor[rgb]{0,0,1}{0.918}          & \textcolor[rgb]{0,0,1}{0.987}          & \textcolor[rgb]{1,0,0}{0.998}          & \textcolor[rgb]{0,0,1}{1.57}          \\
$\text{PEPNet}_{\text{tiny}}$\cite{ren2024simple}    & 180 x 240  & P              & \textcolor[rgb]{1,0,0}{0.054}         & \textcolor[rgb]{1,0,0}{16.25}         & 0.786          & 0.945          & 0.995          & 2.2           \\
$\text{PointMLP}_{\text{elite}}$\cite{ma2022rethinking} & 180 x 240  & P              & 0.68          & 924           & 0.840          & 0.977          & \textcolor[rgb]{0,0,1}{0.997}          & 1.96          \\
PointNet++\cite{qi2017pointnet++}      & 180 x 240  & P              & 1.46          & 1099          & 0.607          & 0.866          & 0.988          & 3.02          \\
PointNet\cite{qi2017pointnet}        & 180 x 240  & P              & 3.46          & 450           & 0.322          & 0.596          & 0.896          & 5.18          \\
CNN\cite{chen20233et}              & 60 x 80    & F              & 0.40          & \textcolor[rgb]{0,0,1}{18.4}          & 0.578          & 0.774          & 0.914          & -             \\
CB-ConvLSTM\cite{chen20233et}    & 60 x 80    & F              & 0.42          & 18.68         & 0.889          & 0.971          & 0.995          & -             \\
ConvLSTM\cite{chen20233et}         & 60 x 80    & F              & 0.42          & 42.61         & 0.887          & 0.971          & 0.994          & -             \\ \hline
\end{tabular}
}
\caption{The results of FAPNet on synthetic SEET dataset. The best results are marked in red and the suboptimal ones are marked in blue. If the frame-based method changes the resolution to 180x240, then the ConvLSTM's computational resource is 1.65 M and 560 M FLOPs.}
\label{table: outdoor}
\end{table*}

\begin{figure*}[ht]
  \centering
   \includegraphics[width=0.9\linewidth]{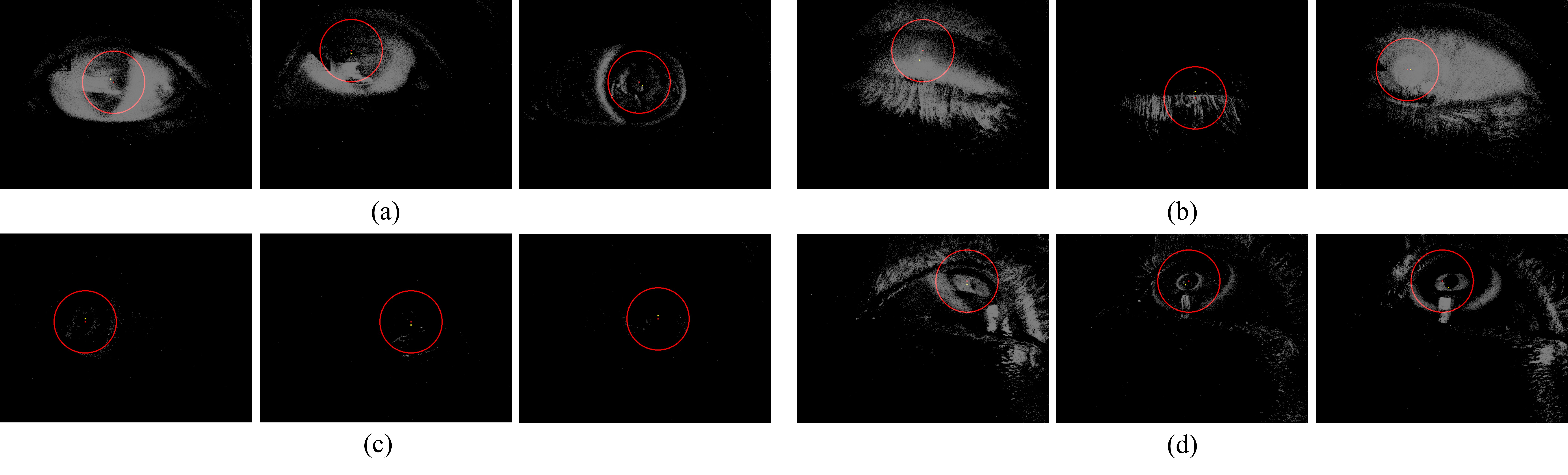}

   \caption{Visualization of the predicted pupil location in various scenarios. (a) is the case contains a large number of events. (b) is the case that existing lots of noise events, such as eyelashes. (c) is the scenario with a few events. (d) is the cases with different pitch angles of eyes.}
   \label{fig:output} 
\end{figure*}

\subsection{Loss Function}
The weighted Mean Squared Error (WMSE) is selected as the loss function to constrain the distance between the predicted pupil location and the ground truth label to guide the network training. The loss function is shown as follows:
\begin{align}
\label{eq:wmse}
\mathcal{L} = \nonumber w_x \cdot \frac{1}{n} \sum_{i=1}^{n} (x_{\text{pred}, i} - x_{\text{label}, i})^2  & \\ + w_y \cdot \frac{1}{n} \sum_{i=1}^{n} (y_{\text{pred}, i} - y_{\text{label}, i})^2,
\end{align}
where $x_{pred}$ and $y_{pred}$ stand for the prediction value along the x and y dimensions. $x_{label}$ and $y_{label}$ are the ground truth value along the x and y dimensions. $w_x$ and $w_y$ are the hyperparameters to adjust the weight of the x dimension and y dimension in the loss function. $i$ means the $i_{th}$ pixel of the prediction or ground truth label.


\section{Experiments}
In experiments, we evaluate the proposed method in various datasets, including a real event dataset provided by the challenge and a synthetic event dataset. The performance of the eye tracking method is measured by the accuracy of pupil location prediction. The Euclidean distance between the predicted pupil location and the label is selected as the objective evaluation metric. Distance error within $p$ pixels is regarded as successful in the estimation of the tracking rate. In the eye tracking challenge, $p_{10}$ is chosen as the tracking score in the leaderboard. 
\subsection{Implement Details}

Our server leverages the PyTorch deep learning framework and selects the AdamW optimizer with an initial learning rate set to $1\cdot e^{-3}$, which is reduced at the 100th and 120th epochs, accompanied by a weight decay parameter of $1\cdot e^{-4}$. This configuration is meticulously chosen to enhance the model's convergence and performance through adaptive learning rate adjustments. Training is conducted on an NVIDIA GeForce RTX 4090 GPU with 24GB of memory, enabling a batch size of 256.

\subsection{Network Structure}
In the challenge, we conducted experiments using vanilla PEPNet. For the synthetic SEET dataset, we further streamlined the entire architecture, incorporating frequency adaptive mechanism and Inter Sample LSTM module to develop a novel architecture, FAPNet. The MLPs’ dimensions in the extractor structure are set to [64, 128, 256] and [32, 64, 128] in PEPNet and FAPNet, respectively. The Bi-LSTM hidden layer dimension is set to 128 and 64 in the Inter Group Aggregation of two models. Specifically, FAPNet uses a sequence length $S$ of 20 and an Inter Sample LSTM with a hidden layer dimension of 128.
\begin{figure*}[ht]
  \centering
   \includegraphics[width=0.8\linewidth]{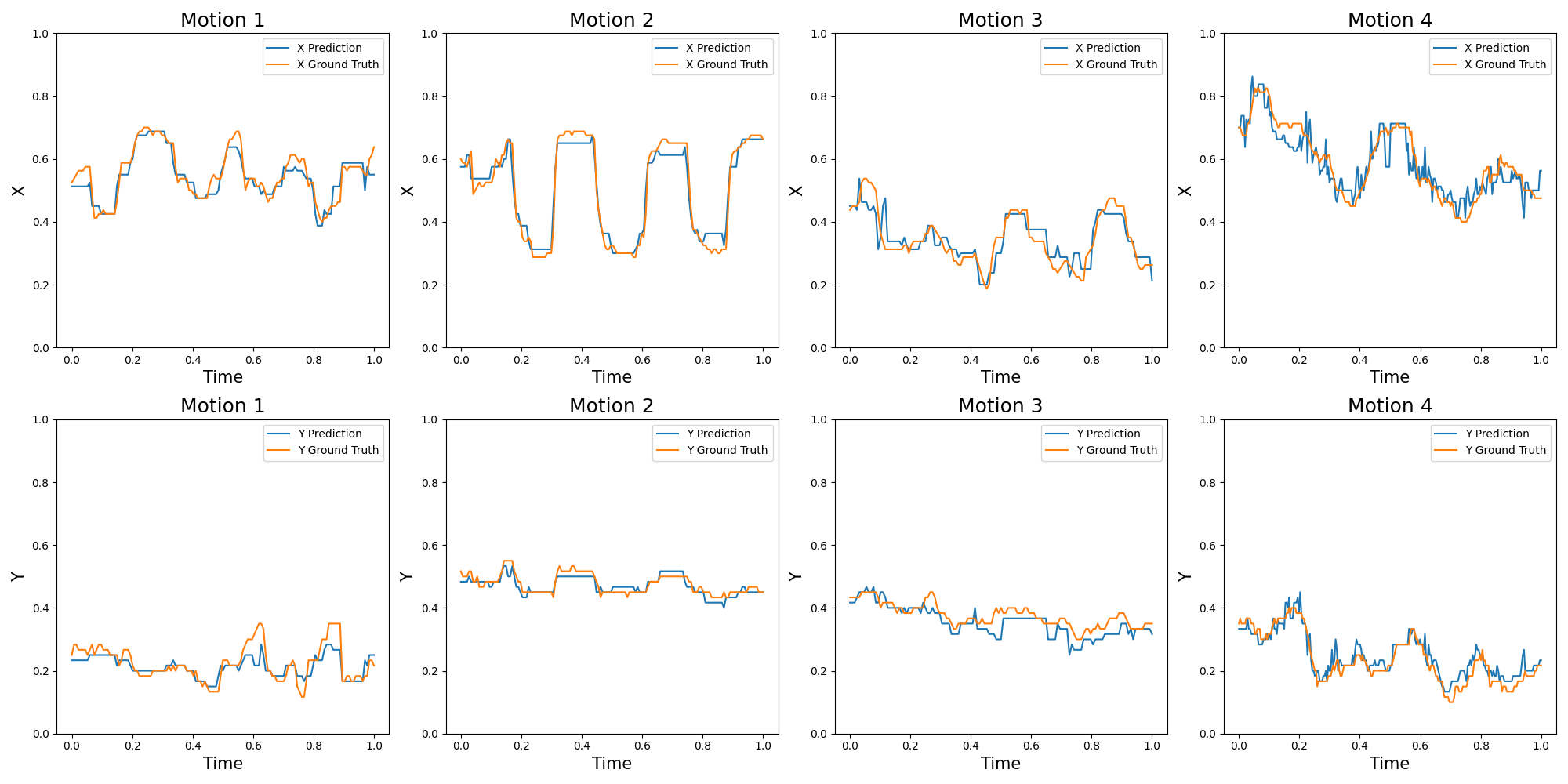}

   \caption{Motion trajectory comparison. The first row is the x-coordinate trajectory along the temporal dimension. The second row is the y-coordinate trajectory of each motion. The x-axis and y-axis are normalized for a better visualization.}
   \label{fig:trajectory}
\end{figure*}

\begin{figure*}[ht]
  \centering
   \includegraphics[width=0.8\linewidth]{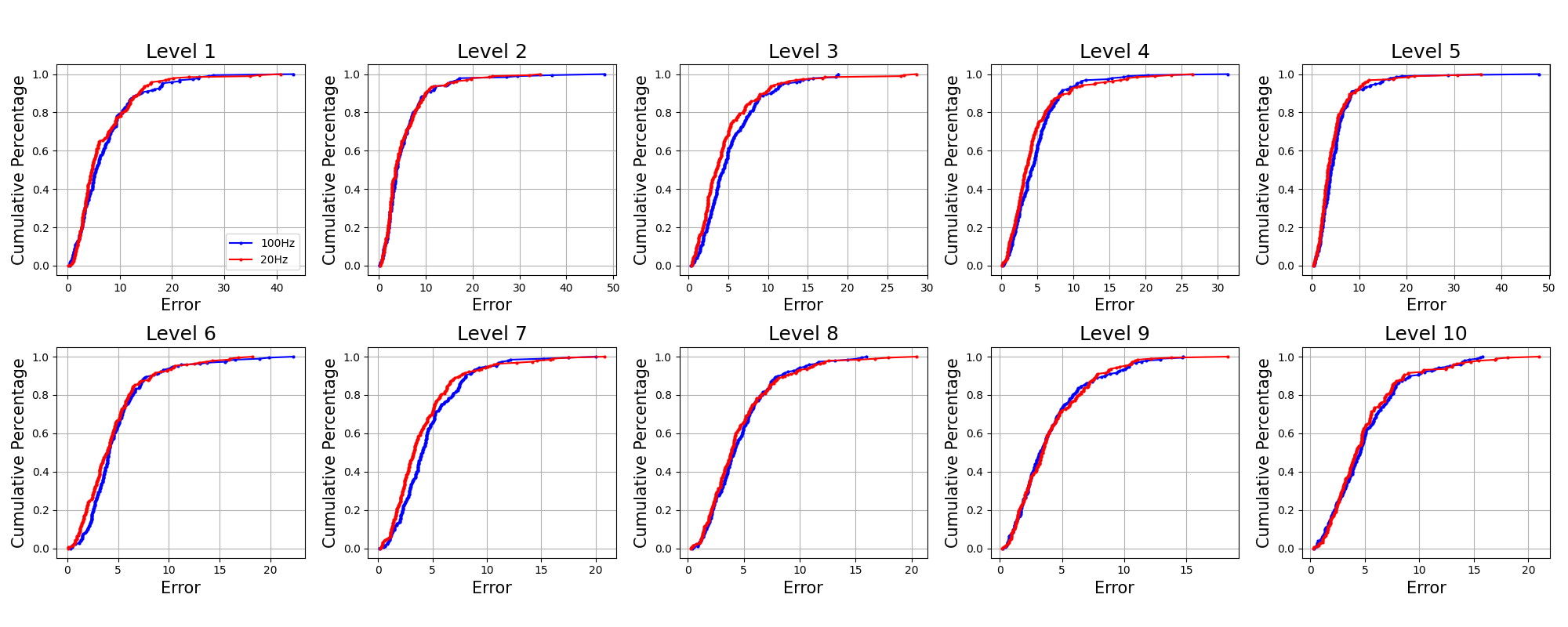}

   \caption{Frequency adaptive window analysis. The testing samples are divided into ten levels according to the event numbers. The x-axis is the distance error between the ground truth label and the prediction, the y-axis is the cumulative percentage of the error. A higher cumulative percentage with the same error stands for better predicting accuracy.}
   \label{fig:100vs20}
\end{figure*}

\subsection{Synthetic Event Data Results}

To demonstrate the effective performance of FAPNet while maintaining low computational cost, we conduct rich comparison experiments among various Point Cloud Networks \cite{qi2017pointnet,qi2017pointnet++,ma2022rethinking,ren2024simple} and other eye tracking methods. The Parameters and FLOPs are selected as the metrics of computational cost. $p_{3}$, $p_{5}$, $p_{10}$ and the mse metrics are chosen for the tracking accuracy measurement.

\cref{table: outdoor} illustrates that the optimal outcomes are highlighted in bold, whereas the second-best solutions are denoted in blue. In comparison to classic point cloud networks, such as PointNet, PointNet++ and $\text{PointMLP}_{\text{elite}}$, FAPNet stands out by significantly enhancing accuracy and reducing distance error while consuming fewer computational resources. Specifically, FAPNet boosts the $p_3$ metric by 59.8\% compared to PointNet, 31.7\% relative to PointNet++ and 8\% in contrast to $\text{PointMLP}_{\text{elite}}$, while requiring fewer than 20 times the FLOPs. 

CNN, ConvLSTM and CB-ConvLSTM eye tracking methods \cite{chen20233et} are selected for comparison. FAPNet exhibits a remarkable improvement in tracking accuracy, with a 3\% increase in the $p_3$ and a 2\% rise in the $p_5$, while maintaining a computational cost comparable to the aforementioned methods. Besides, compared to these frame-based methods, the parameters and FLOPs in FAPNet remain constant regardless of the input camera's resolution. That enables FAPNet to apply in different resolution edge devices. 

As shown in \cref{table: outdoor}, FAPNet is a lightweight, upgraded version of PEPNet\cite{ren2024simple} that achieves comparable accuracy in all metrics with significantly less computation, almost one-tenth of the original. Specifically, FAPNet incorporates a frequency adaptive mechanism to enhance tracking accuracy across varying pupil movement speeds. The Inter Sample LSTM effectively leverages the temporal correlations of the sequential samples, resulting in superior tracking performance. Compared to the vanilla PEPNet, FAPNet network is more lightweight, making it better suited for application on resource-constrained edge devices.
\subsection{Challenge Results}
In the Event-based Eye Tracking Challenge, the vanilla PEPNet is utilized as the eye tracking network. It can achieve $p_{3}$ accuracy at 49.08\%, $p_{5}$ at 80.67\% and $p_{10}$ at 97.95\%, with a mean Euclidean distance of 3.51.

The visualized tracking result is demonstrated in Fig. \ref{fig:output}. As illustrated in Section \ref{sec:Data}, the real event dataset contains rich diversity. In Fig. \ref{fig:output}, we visualize several cases in the tracking motions to demonstrate the robust and high accuracy tracking of the proposed method. Each figure is generated by accumulating the events in a 50ms time interval as an event frame, which contains a red dot means the ground truth label of the pupil location and a yellow dot means the predicted pupil location. The red circle in each event frame highlights the distance between the ground truth label and the prediction within 10 pixels in the $80 \times 60$ resolution, which stands for the range of $p_{10}$ accuracy.

In \cref{fig:output}, we select four typical scenarios for demonstration. \cref{fig:output} (a) shows the cases with a large number of events correlated with the eye movement during the time interval. The sample with high correlation events can provide rich information on the eye movement. During the eye movement, exits noise events affect the final tracking accuracy. As shown in \cref{fig:output} (b), the proposed method is robust enough to regress the pupil location with high accuracy when the eyes are closed or disturbed by eyelashes. \cref{fig:output} (c) shows the performance of the proposed method in the cases with few events, the time window expanding operation enables impressive tracking accuracy in lack-of-event cases. In \cref{fig:output} (d), the pitch angles are various in the eye motions. Combining the information Aggregation, the sparse spatial and temporal information can be fully utilized for the high accurate pupil location prediction. 

The proposed eye tracking method achieves high accuracy tracking for eye movement by predicting the pupil location of each 50ms sample. Fig. \ref{fig:output} demonstrates the robust pupil location regression of each sample, the tracking accuracy is shown in Fig. \ref{fig:trajectory}. For better visualization, the trajectory of the eye movement is separated into the trajectory of the x coordinate and y coordinate along the temporal dimension. The comparison between the ground truth and the predicted trajectory reflects the promising tracking performance of the proposed tracking method.

\subsection{Discussion}

\begin{figure}[ht]
  \centering
   \includegraphics[width=0.85\linewidth]{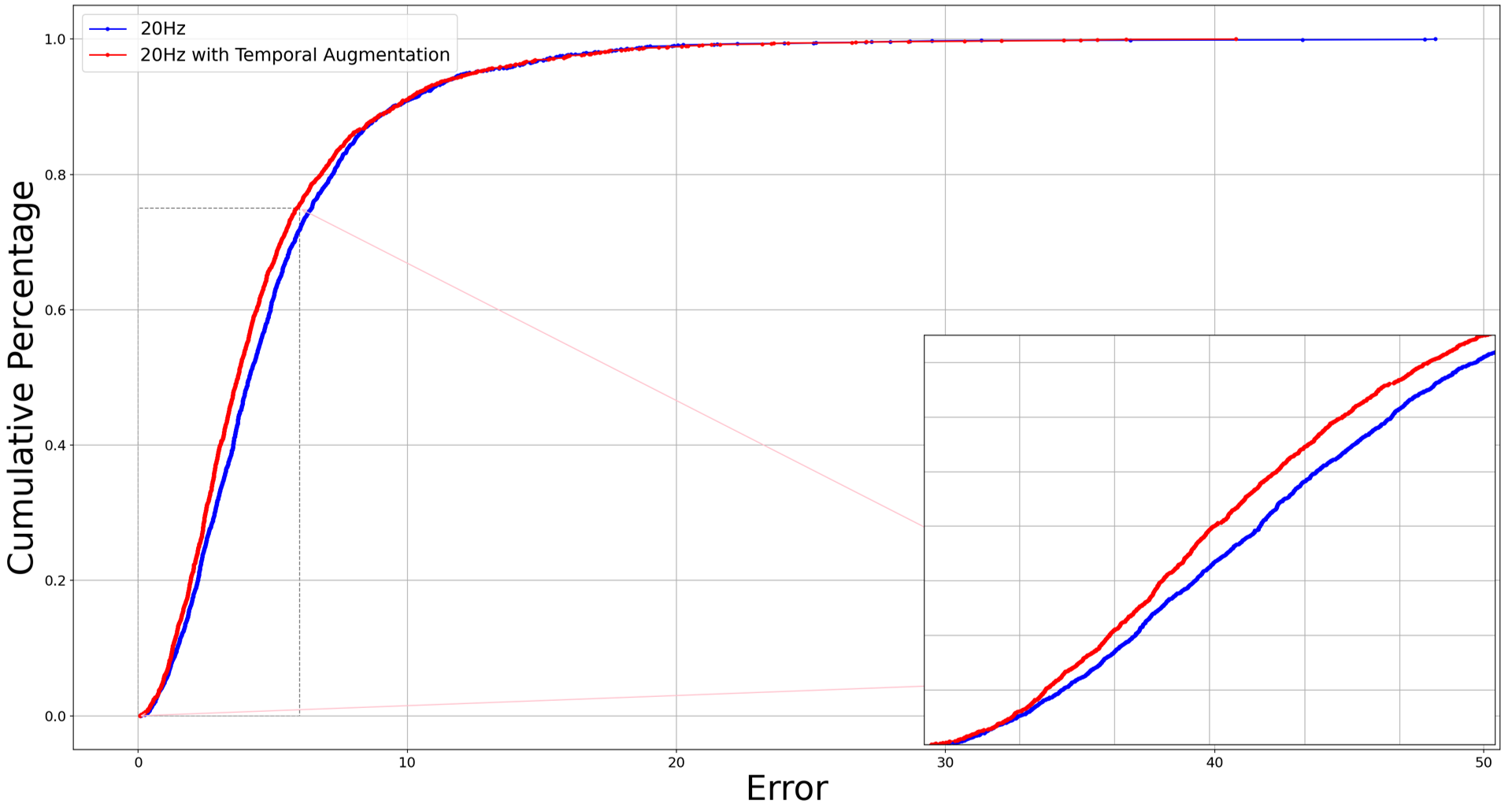}

   \caption{Trajectory augmentation analysis. The x axis is the distance error between the ground truth label and the prediction. The y axis is the cumulative error of the error. The right bottom plot is zooming in the cumulative percentage correlated to the error in the 0-10 range for better comparison.}
   \label{fig:20vs20t}
\end{figure}

To analyze the necessity of frequency adaptive mechanism, we conduct comparison experiments between the samples with different time window sizes. As can be seen in Fig. \ref{fig:100vs20}, the testing samples are divided into ten levels according to the event numbers in each sample. In each sub-figure, the x axis is the distance error between the prediction and the ground truth, the y axis is the cumulative percentage of the error. A higher cumulative percentage with the same error value stands for better predicting accuracy. In Fig. \ref{fig:100vs20}, the red line stands for the results of samples with a 10ms window size and the blue line is the results of samples with a 50ms window size. It can be found that the distance error is decreased when the sample with larger time window size. 

The frequency adaptive mechanism enlarges the time window size when the events are limited. The comparison experiment between 100Hz and 20Hz in Fig. \ref{fig:100vs20} can be regarded as an extreme situation, which enlarges the time window of all samples. The comparison results show that the frequency adaptive mechanism can decrease the distance error by expanding the time window of the samples with limited events. When the sample with large number of events which is highly correlated with pupil movement, the tracking frequency can be switched to a high level. The frequency adaptive mechanism adjusts the tracking frequency according to the frequency of the pupil movement. It enables efficient eye tracking in various scenarios.


In Fig. \ref{fig:20vs20t}, we analysis the effectiveness of the trajectory augmentation method. The cumulative distribution of errors demonstrate that the model trained with the trajectory augmentation can achieve better tracking accuracy with the reducing distance error. It can avoid the model over-fitting to specific trajectory and have stronger generalization ability, which is important to achieve high accuracy eye tracking in the practical applications. 
\section{Conclusion}

In this paper, we present FAPNet, a lightweight eye-tracking model that directly processes the event cloud. It achieves comparable SOTA accuracy while substantially reducing computational requirements. FAPNet consumes low computational resources regardless of the resolution of the sensors, making it ideal for deployment on resource-constrained edge devices and advanced hardware architecture \cite{fu2024ds,liu2024afpr}.

\textbf{Acknowledgment.} This work was supported in part by the Young Scientists Fund of the National Natural Science Foundation of China (Grant 62305278), as well as the Hong Kong University of Science and Technology (Guangzhou) Joint Funding Program under Grant 2023A03J0154 and 2024A03J0618.

{
    \small
    \bibliographystyle{unsrt}
    \bibliography{main}

\begin{thebibliography}{10}

\bibitem{guenter2012foveated}
Brian Guenter, Mark Finch, Steven Drucker, Desney Tan, and John Snyder.
\newblock Foveated 3d graphics.
\newblock {\em ACM transactions on Graphics (tOG)}, 31(6):1--10, 2012.

\bibitem{konrad2020gaze}
Robert Konrad, Anastasios Angelopoulos, and Gordon Wetzstein.
\newblock Gaze-contingent ocular parallax rendering for virtual reality.
\newblock {\em ACM Transactions on Graphics (TOG)}, 39(2):1--12, 2020.

\bibitem{ramachandran2012encyclopedia}
Vilayanur~S Ramachandran.
\newblock {\em Encyclopedia of human behavior}.
\newblock Academic Press, 2012.

\bibitem{angelopoulos2021event}
Anastasios~N Angelopoulos, Julien~NP Martel, Amit~P Kohli, J{\"o}rg Conradt, and Gordon Wetzstein.
\newblock Event-based near-eye gaze tracking beyond 10,000 hz.
\newblock {\em IEEE Transactions on Visualization and Computer Graphics}, 27(5):2577--2586, 2021.

\bibitem{chen20233et}
Qinyu Chen, Zuowen Wang, Shih-Chii Liu, and Chang Gao.
\newblock 3et: Efficient event-based eye tracking using a change-based convlstm network.
\newblock In {\em 2023 IEEE Biomedical Circuits and Systems Conference (BioCAS)}, pages 1--5. IEEE, 2023.

\bibitem{yang2021edge}
Yuchao Yang, Hongwei Ren, Chenghao Li, Chenchen Ding, and Hao Yu.
\newblock An edge-device based fast fall detection using spatio-temporal optical flow model.
\newblock In {\em 2021 43rd Annual International Conference of the IEEE Engineering in Medicine \& Biology Society (EMBC)}, pages 5067--5071. IEEE, 2021.

\bibitem{lichtensteiner2008128x128}
Patrick Lichtensteiner, Christoph Posch, and T~Delbruck.
\newblock A 128x128 120db 15$\mu$s latency asynchronous temporal contrast vision sensor.
\newblock {\em IEEE Journal of Solid-State Circuits}, (2):566--576, 2008.

\bibitem{rebecq2019high}
Henri Rebecq, Ren{\'e} Ranftl, Vladlen Koltun, and Davide Scaramuzza.
\newblock High speed and high dynamic range video with an event camera.
\newblock {\em IEEE transactions on pattern analysis and machine intelligence}, 43(6):1964--1980, 2019.

\bibitem{zhang2022unifying}
Xiang Zhang and Lei Yu.
\newblock Unifying motion deblurring and frame interpolation with events.
\newblock In {\em Proceedings of the IEEE/CVF Conference on Computer Vision and Pattern Recognition}, pages 17765--17774, 2022.

\bibitem{jiang2020learning}
Zhe Jiang, Yu~Zhang, Dongqing Zou, Jimmy Ren, Jiancheng Lv, and Yebin Liu.
\newblock Learning event-based motion deblurring.
\newblock In {\em Proceedings of the IEEE/CVF Conference on Computer Vision and Pattern Recognition}, pages 3320--3329, 2020.

\bibitem{stoffregen2022event}
Timo Stoffregen, Hossein Daraei, Clare Robinson, and Alexander Fix.
\newblock Event-based kilohertz eye tracking using coded differential lighting.
\newblock In {\em Proceedings of the IEEE/CVF Winter Conference on Applications of Computer Vision}, pages 2515--2523, 2022.

\bibitem{zhao2024ev}
Guangrong Zhao, Yurun Yang, Jingwei Liu, Ning Chen, Yiran Shen, Hongkai Wen, and Guohao Lan.
\newblock Ev-eye: Rethinking high-frequency eye tracking through the lenses of event cameras.
\newblock {\em Advances in Neural Information Processing Systems}, 36, 2024.

\bibitem{ren2024simple}
Hongwei Ren, Jiadong Zhu, Yue Zhou, Haotian Fu, Yulong Huang, and Bojun Cheng.
\newblock A simple and effective point-based network for event camera 6-dofs pose relocalization.
\newblock {\em arXiv preprint arXiv:2403.19412}, 2024.

\bibitem{innocenti2021temporal}
Simone~Undri Innocenti, Federico Becattini, Federico Pernici, and Alberto Del~Bimbo.
\newblock Temporal binary representation for event-based action recognition.
\newblock In {\em 2020 25th International Conference on Pattern Recognition (ICPR)}, pages 10426--10432. IEEE, 2021.

\bibitem{bi2020graph}
Yin Bi, Aaron Chadha, Alhabib Abbas, Eirina Bourtsoulatze, and Yiannis Andreopoulos.
\newblock Graph-based spatio-temporal feature learning for neuromorphic vision sensing.
\newblock {\em IEEE Transactions on Image Processing}, 29:9084--9098, 2020.

\bibitem{huang2024clif}
Yulong Huang, Xiaopeng Lin, Hongwei Ren, Yue Zhou, Zunchang Liu, Haotian Fu, Biao Pan, and Bojun Cheng.
\newblock Clif: Complementary leaky integrate-and-fire neuron for spiking neural networks.
\newblock {\em arXiv preprint arXiv:2402.04663}, 2024.

\bibitem{chamorro2022event}
William Chamorro, Joan Sola, and Juan Andrade-Cetto.
\newblock Event-based line slam in real-time.
\newblock {\em IEEE Robotics and Automation Letters}, 7(3):8146--8153, 2022.

\bibitem{krafka2016eye}
Kyle Krafka, Aditya Khosla, Petr Kellnhofer, Harini Kannan, Suchendra Bhandarkar, Wojciech Matusik, and Antonio Torralba.
\newblock Eye tracking for everyone.
\newblock In {\em Proceedings of the IEEE conference on computer vision and pattern recognition}, pages 2176--2184, 2016.

\bibitem{bardow2016simultaneous}
Patrick Bardow, Andrew~J Davison, and Stefan Leutenegger.
\newblock Simultaneous optical flow and intensity estimation from an event camera.
\newblock In {\em Proceedings of the IEEE conference on computer vision and pattern recognition}, pages 884--892, 2016.

\bibitem{zihao2018unsupervised}
Alex Zihao~Zhu, Liangzhe Yuan, Kenneth Chaney, and Kostas Daniilidis.
\newblock Unsupervised event-based optical flow using motion compensation.
\newblock In {\em Proceedings of the European Conference on Computer Vision (ECCV) Workshops}, pages 0--0, 2018.

\bibitem{wang2019space}
Qinyi Wang, Yexin Zhang, Junsong Yuan, and Yilong Lu.
\newblock Space-time event clouds for gesture recognition: From rgb cameras to event cameras.
\newblock In {\em 2019 IEEE Winter Conference on Applications of Computer Vision (WACV)}, pages 1826--1835. IEEE, 2019.

\bibitem{ren2023ttpoint}
Hongwei Ren, Yue Zhou, Haotian Fu, Yulong Huang, Renjing Xu, and Bojun Cheng.
\newblock Ttpoint: A tensorized point cloud network for lightweight action recognition with event cameras.
\newblock In {\em Proceedings of the 31st ACM International Conference on Multimedia}, pages 8026--8034, 2023.

\bibitem{planamente2021da4event}
Mirco Planamente, Chiara Plizzari, Marco Cannici, Marco Ciccone, Francesco Strada, Andrea Bottino, Matteo Matteucci, and Barbara Caputo.
\newblock Da4event: towards bridging the sim-to-real gap for event cameras using domain adaptation.
\newblock {\em IEEE Robotics and Automation Letters}, 6(4):6616--6623, 2021.

\bibitem{ren2023spikepoint}
Hongwei Ren, Yue Zhou, FU~Haotian, Yulong Huang, LIN Xiaopeng, Jie Song, and Bojun Cheng.
\newblock Spikepoint: An efficient point-based spiking neural network for event cameras action recognition.
\newblock In {\em The Twelfth International Conference on Learning Representations}, 2023.

\bibitem{ren2024rethinking}
Hongwei Ren, Yue Zhou, Jiadong Zhu, Haotian Fu, Yulong Huang, Xiaopeng Lin, Yuetong Fang, Fei Ma, Hao Yu, and Bojun Cheng.
\newblock Rethinking efficient and effective point-based networks for event camera classification and regression: Eventmamba.
\newblock {\em arXiv preprint arXiv:2405.06116}, 2024.

\bibitem{qi2017pointnet}
Charles~R Qi, Hao Su, Kaichun Mo, and Leonidas~J Guibas.
\newblock Pointnet: Deep learning on point sets for 3d classification and segmentation.
\newblock In {\em Proceedings of the IEEE conference on computer vision and pattern recognition}, pages 652--660, 2017.

\bibitem{qi2017pointnet++}
Charles~Ruizhongtai Qi, Li~Yi, Hao Su, and Leonidas~J Guibas.
\newblock Pointnet++: Deep hierarchical feature learning on point sets in a metric space.
\newblock {\em Advances in neural information processing systems}, 30, 2017.

\bibitem{wu2019pointconv}
Wenxuan Wu, Zhongang Qi, and Li~Fuxin.
\newblock Pointconv: Deep convolutional networks on 3d point clouds.
\newblock In {\em Proceedings of the IEEE/CVF Conference on computer vision and pattern recognition}, pages 9621--9630, 2019.

\bibitem{zhao2021point}
Hengshuang Zhao, Li~Jiang, Jiaya Jia, Philip~HS Torr, and Vladlen Koltun.
\newblock Point transformer.
\newblock In {\em Proceedings of the IEEE/CVF international conference on computer vision}, pages 16259--16268, 2021.

\bibitem{ma2022rethinking}
Xu~Ma, Can Qin, Haoxuan You, Haoxi Ran, and Yun Fu.
\newblock Rethinking network design and local geometry in point cloud: A simple residual mlp framework.
\newblock {\em arXiv preprint arXiv:2202.07123}, 2022.

\bibitem{qian2022pointnext}
Guocheng Qian, Yuchen Li, Houwen Peng, Jinjie Mai, Hasan Hammoud, Mohamed Elhoseiny, and Bernard Ghanem.
\newblock Pointnext: Revisiting pointnet++ with improved training and scaling strategies.
\newblock {\em Advances in Neural Information Processing Systems}, 35:23192--23204, 2022.

\bibitem{xiao2023unsupervised}
Aoran Xiao, Jiaxing Huang, Dayan Guan, Xiaoqin Zhang, Shijian Lu, and Ling Shao.
\newblock Unsupervised point cloud representation learning with deep neural networks: A survey.
\newblock {\em IEEE Transactions on Pattern Analysis and Machine Intelligence}, 2023.

\bibitem{yang2019modeling}
Jiancheng Yang, Qiang Zhang, Bingbing Ni, Linguo Li, Jinxian Liu, Mengdie Zhou, and Qi~Tian.
\newblock Modeling point clouds with self-attention and gumbel subset sampling.
\newblock In {\em Proceedings of the IEEE/CVF conference on computer vision and pattern recognition}, pages 3323--3332, 2019.

\bibitem{morimoto2005eye}
Carlos~H Morimoto and Marcio~RM Mimica.
\newblock Eye gaze tracking techniques for interactive applications.
\newblock {\em Computer vision and image understanding}, 98(1):4--24, 2005.

\bibitem{wang2017real}
Kang Wang and Qiang Ji.
\newblock Real time eye gaze tracking with 3d deformable eye-face model.
\newblock In {\em Proceedings of the IEEE International Conference on Computer Vision}, pages 1003--1011, 2017.

\bibitem{guestrin2006general}
Elias~Daniel Guestrin and Moshe Eizenman.
\newblock General theory of remote gaze estimation using the pupil center and corneal reflections.
\newblock {\em IEEE Transactions on biomedical engineering}, 53(6):1124--1133, 2006.

\bibitem{lu2014adaptive}
Feng Lu, Yusuke Sugano, Takahiro Okabe, and Yoichi Sato.
\newblock Adaptive linear regression for appearance-based gaze estimation.
\newblock {\em IEEE transactions on pattern analysis and machine intelligence}, 36(10):2033--2046, 2014.

\bibitem{wood2016learning}
Erroll Wood, Tadas Baltru{\v{s}}aitis, Louis-Philippe Morency, Peter Robinson, and Andreas Bulling.
\newblock Learning an appearance-based gaze estimator from one million synthesised images.
\newblock In {\em Proceedings of the Ninth Biennial ACM Symposium on Eye Tracking Research \& Applications}, pages 131--138, 2016.

\bibitem{fengreal}
Y~Feng, N~Goulding-Hotta, A~Khan, H~Reyserhove, and Y~Zhu.
\newblock Real-time gaze tracking with event-driven eye segmentation (2022).

\bibitem{ryan2021real}
Cian Ryan, Brian O’Sullivan, Amr Elrasad, Aisling Cahill, Joe Lemley, Paul Kielty, Christoph Posch, and Etienne Perot.
\newblock Real-time face \& eye tracking and blink detection using event cameras.
\newblock {\em Neural Networks}, 141:87--97, 2021.

\bibitem{wang2024ais_event}
Zuowen Wang, Chang Gao, Zongwei Wu, Marcos~V. Conde, Radu Timofte, Shih-Chii Liu, Qinyu Chen, et~al.
\newblock {E}vent-{B}ased {E}ye {T}racking. {AIS} 2024 {C}hallenge {S}urvey.
\newblock In {\em Proceedings of the IEEE/CVF Conference on Computer Vision and Pattern Recognition Workshops}, 2024.

\bibitem{fu2024ds}
Haotian Fu, Yulong Huang, Tingran Chen, Chenyi Fu, Hongwei Ren, Yue Zhou, Shouzhong Peng, Zhirui Zong, Biao Pan, and Bojun Cheng.
\newblock Ds-cim: A 40nm asynchronous dual-spike driven, mram compute-in-memory macro for spiking neural network.
\newblock {\em IEEE Transactions on Circuits and Systems I: Regular Papers}, 2024.

\bibitem{liu2024afpr}
Haobo Liu, Zhengyang Qian, Wei Wu, Hongwei Ren, Zhiwei Liu, and Leibin Ni.
\newblock Afpr-cim: An analog-domain floating-point rram-based compute-in-memory architecture with dynamic range adaptive fp-adc.
\newblock {\em arXiv preprint arXiv:2402.13798}, 2024.

\end{thebibliography}
}


\end{document}